\documentclass[letterpaper, 10 pt, journal, twoside]{IEEEtran}
\ifCLASSINFOpdf
\else
\fi

\usepackage{epsfig}
\usepackage{graphicx}
\usepackage{amsmath}
\usepackage{amssymb}
\usepackage{algorithm}
\usepackage{algpseudocode}
\usepackage{booktabs}
\usepackage{wrapfig}
\usepackage[dvipsnames]{xcolor}
\newcommand{\method}[1]{{\small\textsc{#1}}}
\newcommand{\new}[1]{#1}

\begin{document}
\title{Morphology-Agnostic Visual Robotic Control}

\author{Brian Yang$^{*1}$, Dinesh Jayaraman$^{*1,2}$, Glen Berseth$^1$, Alexei Efros$^1$, and Sergey Levine$^1$%
\thanks{Manuscript received: Sep, 11, 2019; Revised Dec, 2, 2019; Accepted Dec, 10, 2019.}%
\thanks{This paper was recommended for publication by Editor Cesar Cadena upon evaluation of the Associate Editor and Reviewers' comments.
This work was supported by Berkeley DeepDrive, the National Science Foundation under IIS-1700697, the Office of Naval Research, and the DARPA Assured Autonomy Program.} %
\thanks{$^{1}$Berkeley AI Research, UC Berkeley}
\thanks{$^{2}$Facebook AI Research}
\thanks{Digital Object Identifier (DOI): see top of this page.}
}
\markboth{IEEE Robotics and Automation Letters. Preprint Version. Accepted Dec, 2019}
{Yang \MakeLowercase{\textit{et al.}}: Morphology-Agnostic Visual Robotic Control}

\maketitle

\begin{abstract}
Existing approaches for visuomotor robotic control typically require characterizing the robot in advance by calibrating the camera or performing system identification. We propose \method{mavric}, an approach that works with minimal prior knowledge of the robot's morphology, and requires only a camera view containing the robot and its environment and an unknown control interface. \method{mavric} revolves around a mutual information-based method for self-recognition, which discovers visual ``control points'' on the robot body within a few seconds of exploratory interaction, and these control points in turn are then used for visual servoing. \method{mavric} can control robots with imprecise actuation, no proprioceptive feedback, unknown morphologies including novel tools, unknown camera poses, and even unsteady handheld cameras. We demonstrate our method on visually-guided 3D point reaching, trajectory following, and robot-to-robot imitation.
\end{abstract}

\section{Introduction}\label{sec:intro}
\IEEEPARstart{A} child playing an arcade ``claw crane'' game must first visually locate and recognize the claw robot they are controlling, and learn how it responds to various control commands---something they typically accomplish within a few seconds of twiddling the joystick controller. Infants and some animals also exhibit mirror self-recognition, the ability to recognize one's reflection in the mirror as tied to oneself~\cite{gallup1970chimpanzees,asendorpf1996self}. In this paper, we ask: how might a robot perform efficient self-recognition and how might this ability improve robotic control? %

Current robotic control methods typically require precise knowledge of the robot's configuration and kinematics, obtained from accurate geometric models of its body, and joint-level proprioceptive encoders.
As an example, the degrees of freedom of a standard robot arm are fully specified by its joint angles, available through servomotor encoders. 
Given a target pose, a controller can quite easily plan a trajectory of poses, and servo to sequentially reach those poses.

Unfortunately, such proprioception-driven control methods do not generalize to many important settings. What if the robot were made of soft or deformable material, so that its degrees of freedom are not easily enumerated, let alone measured? Even for the rigid robot arm above, introduce, say, a pen into its gripper, and the position of its tip is now unknown and therefore not possible to control. 

Humans easily handle such control tasks by relying on rich sensory feedback, such as from vision, rather than purely on precise proprioception. This makes us less reliant on precisely modeling our bodies and tools, and the richer visual space also enables specifying and performing a larger class of tasks, such as writing with an unfamiliar chalk piece, or catching a ball. %

Existing image-based visual servoing (IBVS)
approaches attempt to bridge this gap by relying on visual feedback to plan towards visual goals, but they require the robot to be pre-specified in some way. For example, they may assume that there exists a reliable detector for a point of interest such as the end-effector of a robot arm, so that its position in the camera view is known. To control a new robot, or the pen in the gripper above, a new detector would have to be provided by a human engineer, either by applying visual markers or by training a visual detector with machine learning. %

We propose \method{mavric}, a self-recognition-enabled approach to IBVS that works ``out of the box'' on arbitrary new or altered robots with no manual specification of any points of interest. %
In a self-recognition phase, the robot locates and characterizes itself through exploratory actions, much like the claw crane game player twiddling their joystick. %
We use simple techniques to accomplish this: a mutual information measure~\cite{shannon1948mathematical} evaluates the responsiveness of various points in the environment, tracked using Lucas-Kanade optical flow computation~\cite{lucas1981iterative}, to the robot control commands. The most responsive points are then exploited in the control phase. \method{mavric} is lightweight, flexible, and fast to adapt, producing responsive ``control points'' for a new robot within a few seconds of interaction.

\begin{figure*}
  \centering
  \includegraphics[width=0.85\linewidth]{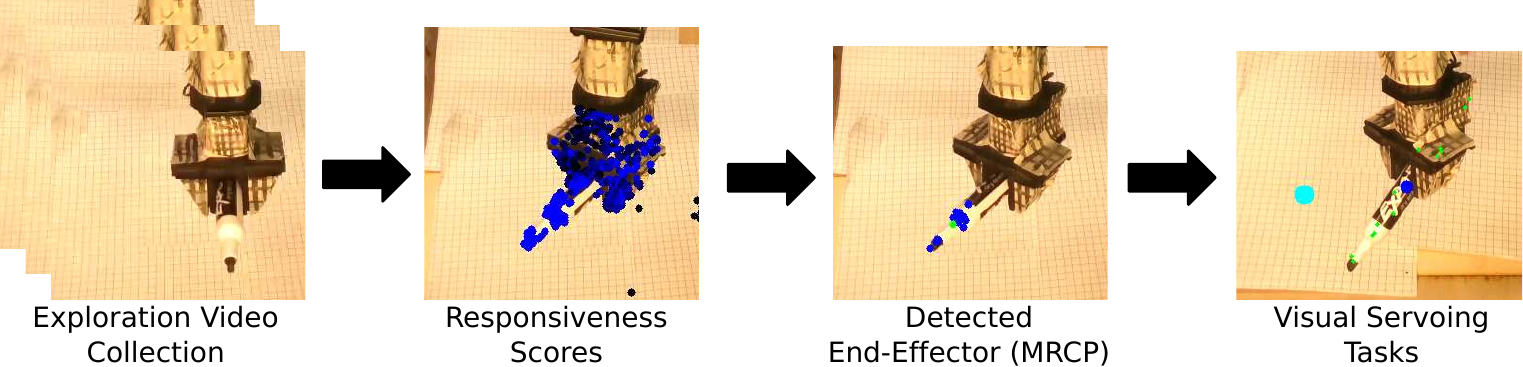}
  \caption{\small{\method{mavric} first collects random exploration video for a few seconds, then computes scores for various tracked points corresponding to their responsiveness to control inputs (shown in blue here). The top-few such points are then averaged (shown as the green point) treated as the end-effector for visual servoing in control tasks.}}
  \label{fig:mavric_method}
\end{figure*}

As we will show, \method{mavric} handles settings that are challenging for today's state-of-the-art robotic control approaches: imprecise actuation, unknown robot morphologies,  unknown camera poses, novel unmodeled tools, and unsteady handheld cameras. %

\section{Related Work}\label{sec:related}

\vspace{0.02in}\noindent\textbf{Image-based visual servoing.~~} Visual servoing methods~\cite{corke1993visual,hutchinson1996tutorial,chaumette2006visual,chaumette2007visual,marchand2005visp} perform feedback control to reduce an error that is measured through camera observations.
Specifically, uncalibrated image-based visual servoing (IBVS) approaches~\cite{hosoda1994versatile,yoshimi1994active,jagersand1997experimental,bonkovi2008population,shademan2010robust,dame2011mutual,lampe2013acquiring,mohta2014vision,bateux2018training,viereck2017learning,andersson1988robot,jang1991concepts,cai20146d,mcfadyen2017image}
typically estimate the differential relationships between control inputs and changes in some visual feature of interest, such as the pixel coordinates of the robot's end-effector in an uncalibrated camera view.
Reinforcement learning approaches have also been proposed for the IBVS setting~\cite{finn2016deep,ebert2017self,sadeghi2017sim2real,levine2018learning,lee2017learning,james2019sim}.

\new{All these IBVS methods produce controllers that are tied to a single robot morphology in some way---for example, they may require visual markers on the robot~\cite{andersson1988robot,jang1991concepts,cai20146d,mcfadyen2017image} or a large dataset of interactions specific to the current robot morphology and environment~\cite{finn2016deep,sadeghi2017sim2real,lee2017learning,james2019sim,shkurti2017underwater,islam2018towards,ebert2017self,levine2018learning}.} In contrast, \method{mavric} performs automatic self-recognition to produce a controller that adapts to new or altered robots within a few seconds.

\vspace{0.02in}\noindent\textbf{Robotic self-recognition.~~} Prior methods have been proposed that learn to recognize the robot's body. %
Michel et al \cite{michel2004motion} use learned characteristic response delays between actions and observed motion, and Natale et al \cite{natale2007sensorimotor} also exploit simple temporal correspondences, relying on periodicity. %
Robot and object keypoints may be discovered by training a keypoint encoder for image reconstruction on a large image dataset~\cite{finn2016deep}, or by weakly supervised multiple-instance learning from a few minutes of video with and without an object of interest~\cite{jonschkowski2018found}. Byravan et al~\cite{byravan2018se3} demonstrate rigid robot link discovery by training a dynamics model with appropriate inductive biases on hundreds of thousands of depth frames annotated with ground truth correspondences.
Closest to \method{mavric}, \cite{edsinger2006can} use a mutual information-based approach to recognize individual links, but different from \method{mavric}, they assume known aspects of the robot's morphology, such as the number of its links and the visual appearance of its body, and the correspondences between action dimensions and the various servos on the robot, and also rely on manually demonstrated robot poses during exploration. Despite these advantages, they report requiring four minutes of exploration, compared to about 20s for \method{mavric}. They also propose a different approach to tracking, which we empirically compare against.

\section{Morphology-Agnostic Visual Control}\label{sec:maverick}\label{sec:setting}

We operate in the following setting: at each time step $t$, a controller has access to raw RGBD image observations from a camera, and the ability to set a $d$-dimensional control input $A(t)$ for the robot. The images contain the robot's body as well as other portions of its environment. We are interested in performing visually guided control tasks, such as reaching and trajectory-following. 

To maximize generality, we make very few assumptions about factors that are commonly treated as fully known in robotic control: (i) We do not know the nature of the robot's embodiment, such as the degrees of freedom, rigidity, or the number and lengths of links in a robot arm. (ii) For the control interface, aside from the standard assumptions made in uncalibrated visual servoing, 
we make one additional assumption that the displacements of points on the robot are a probabilistic function of the control commands. We make no further assumptions about how the controls affect the robot.
We will revisit this point in Sec~\ref{sec:mrcp}. (iii) We do not assume camera calibration.
Our approach, \method{mavric}, works in two phases. In a self-recognition phase (Sec~\ref{sec:mrcp}), it identifies the robot's body and represents it as trackable and characterizable ``control points,'' through unsupervised interaction. In the control phase (Sec~\ref{sec:servoing}), it performs \new{visual servoing to move these control points along desired target trajectories.}

\subsection{Self-Recognition: Robot as Responsive Particles}\label{sec:mrcp}

In the self-recognition phase, we aim to resolve the question: what is the body of the robot? %
In standard robotic control settings, the body is a predefined, physically connected object with rigid links.
Since we do not assume a pre-specified body in our setting, we first define the body in terms of a new notion of ``responsiveness.''

We start by decomposing the observable environment containing the robot into ``particles''--- points in 3D space that may or may not be part of its body. Each particle $P_i$ has an associated position $S_i(t)$ in RGBD camera coordinates ($x$, $y$, depth) from our uncalibrated camera. The task of identifying the body now reduces to assigning a binary label (body / not body) to each such particle.
\vspace{0.05in} \noindent\textbf{Responsiveness.~~}
We define the responsiveness of a particle as the mutual information (MI)~\cite{shannon1948mathematical,cover2012elements} between its motions and the control inputs. Specifically, we execute a random sequence of exploratory control commands $A(t)$ assign a non-negative responsiveness score to each particle $P_i$:
\begin{equation}
R_i \triangleq I(\Delta S_i; A),
\label{eq:responsiveness}
\end{equation}
where $\Delta S_i(t)=S_i(t+1)-S_i(t)$ is the change in position of $P_i$ in response to $A(t)$, and $I(\cdot; \cdot)$ is the MI between two random variables. $R_i$ is zero for points whose motion is completely independent of $A$.

To understand this definition intuitively, note that \mbox{$R_i = I(\Delta S_i; A)=H(\Delta S_i) - H(\Delta S_i | A)$}, where $H(\cdot)$ is entropy and $H(\cdot | \cdot)$ is conditional entropy. High responsiveness corresponds to low conditional entropy $H(\Delta S_i|A)$, \emph{i.e.}, changes in states are predictable given the control inputs. At the same time, the state changes themselves should also be sufficiently expressive, \emph{i.e.}, the unconditional entropy $H(\Delta S_i)$ should be high. A fixed particle in the environment would have zero $H(\Delta S_i | A)$, but also zero H($\Delta S_i$), so it would have zero responsiveness.

Alternatively, since MI is symmetric, $R_i$ may also be written as $H(A) - H(A | \Delta S_i)$. Since $H(A)$ is the same for all particles, high responsiveness corresponds precisely to low conditional entropy $H(A|\Delta S_i)$: control inputs $A$ should be easily recoverable given the state changes $\Delta S_i$. This is consistent with the intuition for discovering ``contingent image regions'' in Atari games used in Choi et al~\cite{choi2018contingencyaware}. 

\begin{figure*}
  \centering
  \includegraphics[width=0.9\textwidth]{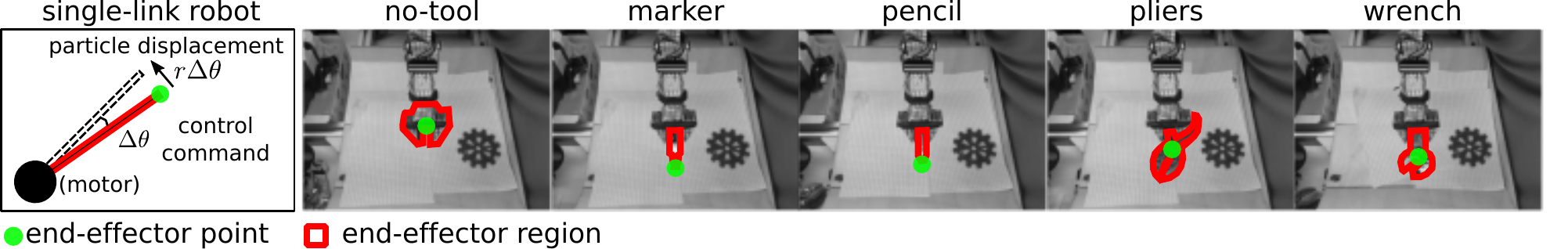}
  \caption{\small{Manual annotations to illustrate the end-effector point (green point) and region (red outline) in different settings: a schematic single link robot (leftmost), followed by our experimental setups with various tools held in a robot arm.}}
  \label{fig:body_def}
\end{figure*}

\vspace{0.05in}
\noindent\textbf{Body and control points.~~} We define the ``body'' $B_\delta$ of a robot as the set of particles whose responsiveness is higher than some threshold $\delta$:
$B_\delta \triangleq \{P_i: R_i>\delta\}$. We call the constituent particles of this set ``control points.''

First, does this definition align with our intuitive notion of a robot's body? Should all points on a robot body be ``responsive'', i.e., do their displacements $\Delta S$ have high MI with the control inputs $A$? For this to hold, $\Delta S$ must be a \emph{probabilistic function} of $A$, i.e., a fixed control input must induce a fixed distribution over $\Delta S$. This is true for velocity control commands as long as the states $S$ explored during the self-recognition phase lie within a small neighborhood. For example, consider a single motor controlling a rigid rod, as in Fig~\ref{fig:body_def} (left). A small angular shift $\Delta\theta$ in the servomotor corresponds to a displacement $r\Delta\theta$ for a particle at a distance $r$ along the rod, in a direction perpendicular to the \emph{current orientation} of the rod. With a significantly different orientation of the rod, the same angular shift would produce a very different displacement.

To account for this, our experiments employ velocity control and a small number of small exploration actions, so that all exploration happens within a small state neighborhood. As we will show empirically, this yields good performance. %

\vspace{0.05in} \noindent \textbf{Most responsive control point (MRCP).~~} We now define the most responsive control point (MRCP) $P^*$ as the particle with the highest responsiveness $R^*=\max_i R_i$. Its position is henceforth denoted $S^*(t)$. In practice, we average the positions of the top-$k$ most responsive particles to compute a robust MRCP. \new{The MRCP is the point that is most responsive to control, which intuitively corresponds to highest maneuverability or dexterity. For this reason, it may be treated as the \emph{end-effector}. %
Fig~\ref{fig:body_def} shows images of various settings from our experiments, with manual annotations of the end effector point and region that we evaluate the MRCP against.}
As an example, consider a hammer held firmly in the gripper of a robotic arm. It is very responsive to the arm's control inputs, and may even contain the MRCP. \new{Replacing the hammer with a loose rope, the rope would be less responsive, and the MRCP would be in the gripper instead.}

\vspace{0.05in} \noindent\textbf{Handling rigid objects.~~} Points on a rigid object, such as a single link of a standard arm, all exhibit the same motion, modulo an invertible affine transformation. Mutual information is known to be invariant under such smooth, invertible mappings, a.k.a.~homeomorphisms (see Kraskov et al \cite{kraskov2004estimating} for a simple proof). This in turn means that points on a rigid object all have the same responsiveness score. For example, for a rigid rod pivoted at one end as in Fig~\ref{fig:body_def}, the midpoint of the rod is just as responsive to any forces applied to the rod as the end of the rod.

To break such ties, we preferentially select points with larger motions. We do this by exploiting the fact that, although mutual information is insensitive to the scale of motion under \emph{infinite} precision, any loss of precision leads naturally to a preference for large motions. We add Gaussian noise to $\Delta S_i$ in Eq.~\ref{eq:responsiveness} to artificially lower the precision and express this preference for large motions.

Note however that it is not always the case that particles that are near the periphery or which move the most are MRCP points. For example, if the last link of a robot arm is broken so that it moves randomly, it might very well have a lower responsiveness score than the previous link. %

\begin{figure*}
    \centering
    \includegraphics[width=0.85\textwidth]{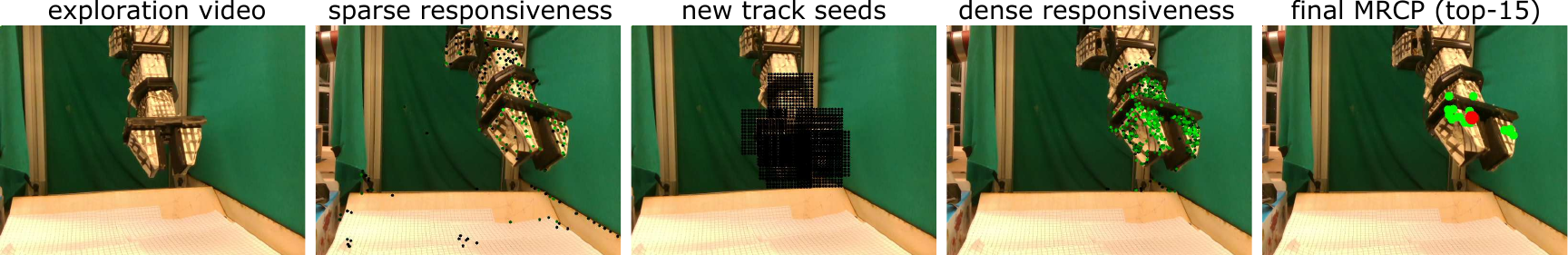}
    \caption{\small{[Best seen in pdf] Self-recognition for MRCP search illustrated on data from a handheld, shaky camera. Frame 1 shows the camera view of the robot during exploration. Frame 2 shows the results of coarse tracking on the exploration video, where the track points overlaid on the image are colored more green if they are more responsive. Frame 3 shows how fine tracking is initialized on the same video around most responsive tracks from the coarse stage, and Frame 4 shows surviving tracks after the fine stage of MRCP search. Frame 5 shows the final result, with the top-15 control points in green and their average position in red. Video in Supp.}} %
    \label{fig:mrcp_handheld}
\end{figure*}

\vspace{0.05in} \noindent\textbf{Implementation details.~~} Fig~\ref{fig:mavric_method} shows a schematic of  the \method{mavric} self-recognition phase. For tracking the positions $S_i(t)$ of particles over time, we use the Lucas-Kanade optical flow estimator~\cite{lucas1981iterative}. %
We only consider points that last the full duration of the exploration actions.
We use the mutual information estimator proposed in~\cite{kraskov2004estimating}, as implemented in~\cite{ver2000non}. In our experiments, control points on a robot arm are discovered within 20 seconds of exploration, sufficient to execute about 100 randomly sampled small actions. Finally, since we are primarily interested in the MRCP for control, we adopt a coarse-to-fine search strategy: In the coarse stage, we initialize point tracking with Shi-Tomasi corner points~\cite{shi1993good}, compute responsiveness scores and select the top-$k$ candidate particles. In the fine stage, we reinitialize tracking with a grid of 15x15 points around each of the selected candidates, and recompute responsiveness scores. \new{The only important hyperparameter in all the above is the noise variance for handling rigid objects. Our experiments in Sec~\ref{sec:mrcp_res} study this and other implementation details such as coarse-to-fine search}. %
Fig~\ref{fig:mrcp_handheld} shows an example of the various stages of MRCP identification with a handheld, shaky camera.

\new{Finally, the Lucas-Kanade tracker often drops particles over time, especially under occlusions. Since \method{mavric} is able to work with very short exploration phases in our experiments, the effect of dropped particles is mitigated. %
However, improved tracking, such as through articulated motion segmentation approaches~\cite{yan2006general}, or by using multiple cameras, may improve robustness.}%

\subsection{Visual Servoing in MRCP Coordinates}\label{sec:servoing}
Once the MRCP is identified, \method{mavric} performs visual servoing for control to transport the MRCP point $S^*$ to a specified goal point $G$. %
This is appropriate for tasks like reaching and pushing, which are normally performed by hand-specified end-effectors in standard control settings.

We use online regression to fit $(A, \Delta S^*)$ tuples to estimate local Jacobian matrices ``on the fly,'' using Broyden updates~\cite{jagersand1997experimental}: $\hat{J}_{t} = \hat{J}_{t-1} + (\Delta S_t - J_{t-1} A_t) A_t^T/\|A_t\|^2$.

The new control input $A_t$ for the current step is then computed \new{quickly and cheaply} using the pseudoinverse of the Jacobian, as $A_t = \eta \hat{J}_t^\dagger (G-S^*(t))$, where $\eta$ is a rate hyperparameter. Once $A_t$ is executed and the new position $S^*$ is measured, the Jacobian matrix is updated as above, and the process repeats until $S^*\approx G$. The Broyden update above is susceptible to noise, since it only uses a single $\Delta S_t$ measurement, hence we apply a batched update comprising the last $T$ tuples of ($A_\tau$, $\Delta S_\tau$) as proposed in~\cite{bonkovi2008population}. In our experiments, we set $T=10$. %

The Jacobian matrix initialization $J_0$ is computed as follows: we start at a random arm position,
sequentially set the control inputs $A_i$ to scaled unit vectors $\epsilon e_i$ along each control dimension, and set the $i$-th column of $J_0$ to the response $\Delta S_i/\epsilon$. In our experiments, we repeat this initialization procedure whenever servoing has failed to get closer to the target in the last 20 steps. %

\vspace{0.05in} \noindent\textbf{Handling tracking failures.~~}  The above discussion of visual servoing depends on reliably continuing to track the MRCP throughout the servoing process. In practice, tracking is imperfect, and the control points are often dropped midway through the task due to occlusions, lighting changes etc. For robustness to such errors, we take the MRCP to be the average of the $K=15$ most responsive control points. If any one point is dropped by the tracker during servoing, the MRCP is set to the average of the remaining points. \new{The larger $K$ is, the greater is the robustness to dropped points, and the lower is the precision in end effector localization since those $K$ points would be spatially more scattered.
In our studies, $K=15$ sufficed for robust MRCP detection, so we study lower $K$ in experiments.} %

\section{Experiments}\label{sec:exp}
We perform experiments using the REPLAB standardized hardware platform~\cite{yang2019replab, yang2019replab2} with an imprecise low-cost manipulator (Trossen WidowX) and an RGB-D camera (Intel Realsense SR300). We evaluate how well \method{mavric}'s self-recognition phase works under varied conditions, and also the overall effectiveness of \method{mavric} for visuomotor servoing tasks. We will release REPLAB-compatible code upon acceptance for reproducibility.

\subsection{Self-Recognition: Discovering End Effectors, Tools, and Robot Morphology}\label{sec:mrcp_res}

First, we evaluate \method{mavric}'s self-recognition phase, described in Sec~\ref{sec:mrcp}. \new{Specifically, we measure its accuracy at locating the robot's end-effector or the tool held in its gripper, with four different tools -- pliers, wrench, marker, pencil, shown in Fig~\ref{fig:body_def}}.

\vspace{0.05in} \noindent \textbf{Simulation.~~} In simulation, we evaluate \method{mavric} with perfect actuation and tracking, as a sanity check. We use a simulated Baxter robot, and perform 100 small exploratory control commands during the self-recognition phase. In our experiments, we precisely identified the end-effector with zero error every single time in this setting. %

\begin{figure*}
    \centering
    \begin{tabular}{ccccc}
    \small{no tool} & \small{pencil} & \small{pliers} & \small{wrench} & \small{amputated} \\
    \includegraphics[width=0.17\textwidth]{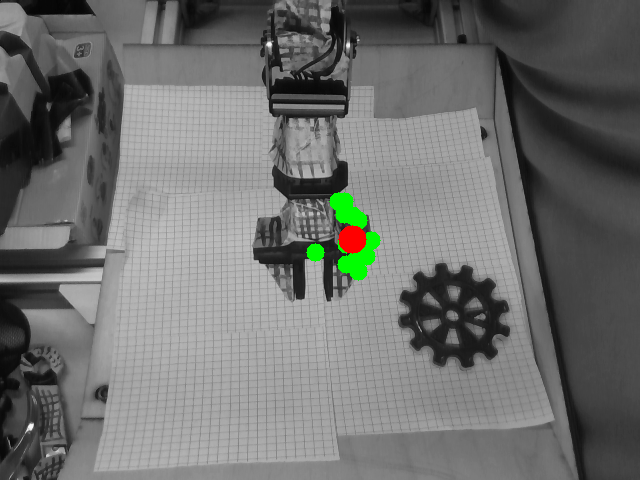} &
    \includegraphics[width=0.17\textwidth]{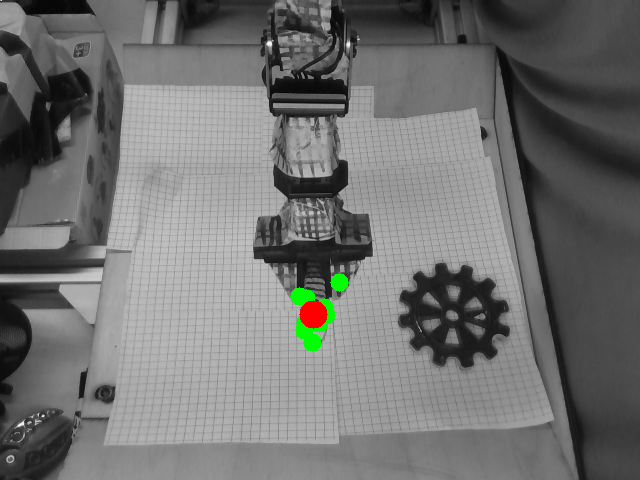} &
    \includegraphics[width=0.17\textwidth]{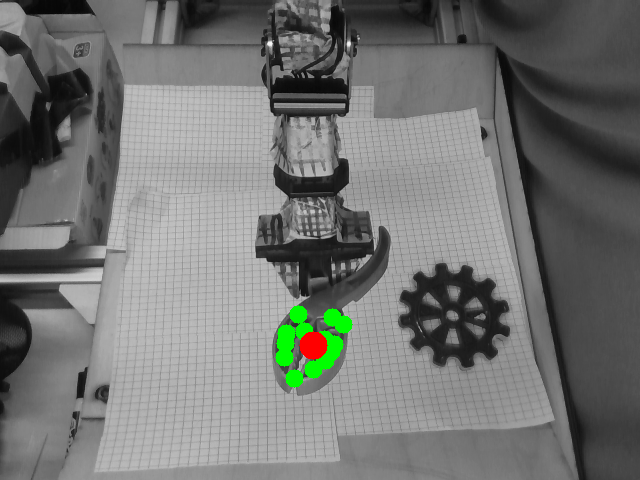} &
    \includegraphics[width=0.17\textwidth]{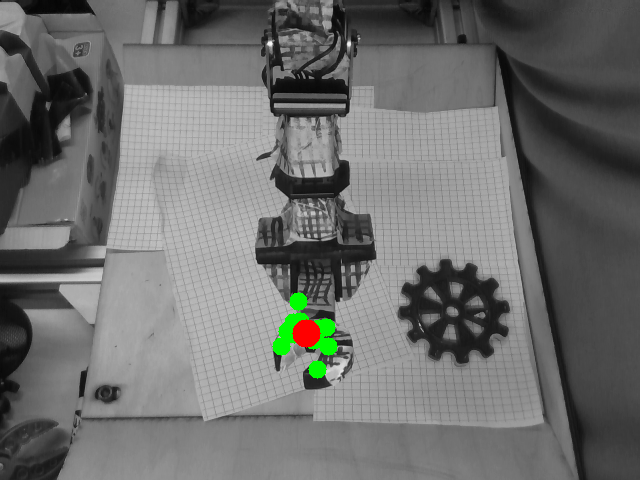} &
    \includegraphics[width=0.17\textwidth]{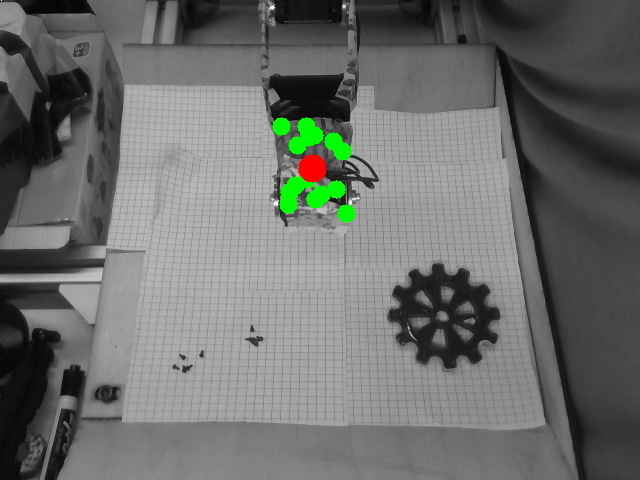}
    \end{tabular}
    \caption{\small{[Best seen in pdf] MRCP identification in various settings. In each setting, the red point is the MRCP, computed as the average of the 15 most responsive points, shown in green. Please see videos in Supp. Fig~\ref{fig:body_def} presents the ground truth end effector annotations used to evaluate these settings.}}
    \label{fig:mrcp_egs}
\end{figure*}

\vspace{0.05in} \noindent \textbf{Real-world experiments.~~}
Next, we evaluate our approach on real-world REPLAB cells with noisy tracking and actuation. In each run, a sequence of 100 random exploratory control commands are executed, which requires about 20 seconds.
Fig~\ref{fig:mrcp_egs} shows some example results of detected control points and MRCP points in each setting.

We quantitatively evaluate how closely \method{mavric}'s MRCP matches the manually annotated ``true end-effector'' of the robot. We annotate an end-effector region as well as a single end effector point in each setting. %
Example annotations are shown in Fig~\ref{fig:body_def}: the end-effector region includes the entire tool or the entire last link of the robot, and the end-effector point is more subjectively chosen for each tool based on how it is typically used, e.g., for the pencil and the marker, we annotate its tip. %
We evaluate several design decisions in \method{mavric}:
1-stage vs 2-stage (coarse-to-fine) MRCP search, values of $K$ for top-K control point selection, and values of noise variance added to point tracks before responsiveness computation. %
No prior work studies self-recognition in as general settings as ours, and the closest approaches~\cite{edsinger2006can,natale2007sensorimotor,michel2004motion} do not have public implementations. In lieu of prior work, we introduce a simple baseline that selects the points that move the largest distance over exploration (``Max-motion''). We compare later against our implementation of Edsinger et al~\cite{edsinger2006can}'s tracking approach.

\begin{figure*}
    \centering
    \includegraphics[width=0.85\textwidth]{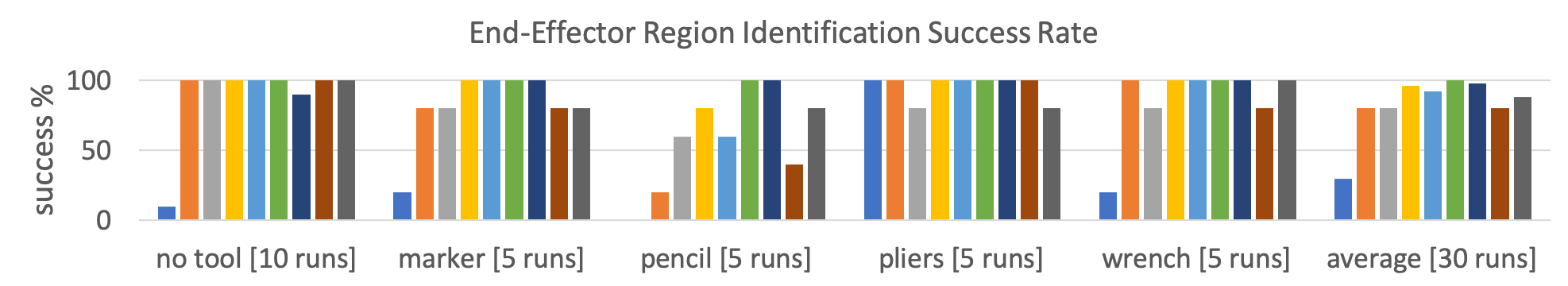} \\
    \includegraphics[width=0.85\textwidth]{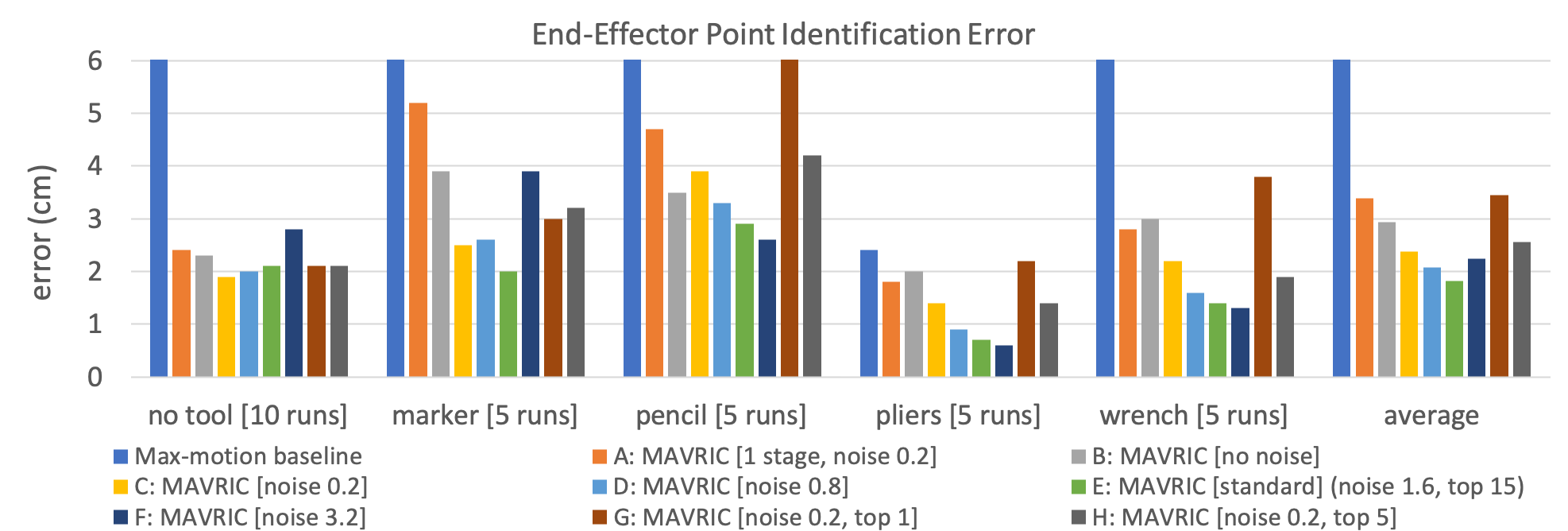}
    \caption{\small{[Best seen in pdf] \textbf{(Top)} End-effector region identification success rate (higher is better), and \textbf{(Bottom)} End-effector identification error (in cm, lower is better)
    in various settings for the maximum-motion baseline, and various ablations of our method. Note that max-motion performs very poorly in most situations (average error 8.8 cm): the error axis is clipped at 6 cm here. Among ablations of \method{mavric}, we study three hyperparameters: number of stages of end-effector ID (default: 2), noise variance (in squared pixel units) before responsiveness computation (default: 1.6), number of top points averaged to compute the MRCP (default: 15).}}
    \label{fig:mrcp_plots}
\end{figure*}

Fig~\ref{fig:mrcp_plots} shows these quantitative results. Max-motion peforms very poorly in nearly all settings, while most variants of \method{mavric} get close to perfect end-effector region identification success rate.
Max-motion has a number of conceptual weakenesses, since it cannot distinguish between \emph{intentional} and \emph{accidental} motion, and is vulnerable to moving distractors. However, even in experiments without such distractors, we find that Max-motion often fails due to sensitivity to noisy sensing and tracking (examples in Supp).

\begin{figure*}

    \centering
    \begin{tabular}{ccccc}
    ~  & \small{variance 0.0}  & \small{variance 0.4} & \small{variance 0.8} & \small{variance 1.6} \\
    \includegraphics[width=0.17\textwidth]{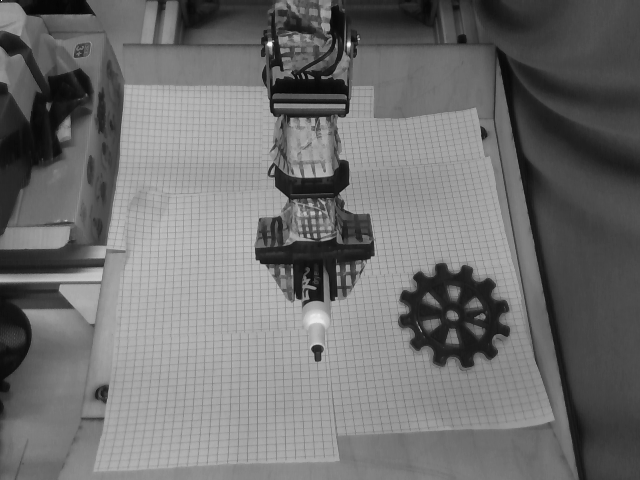} &
    \includegraphics[width=0.17\textwidth]{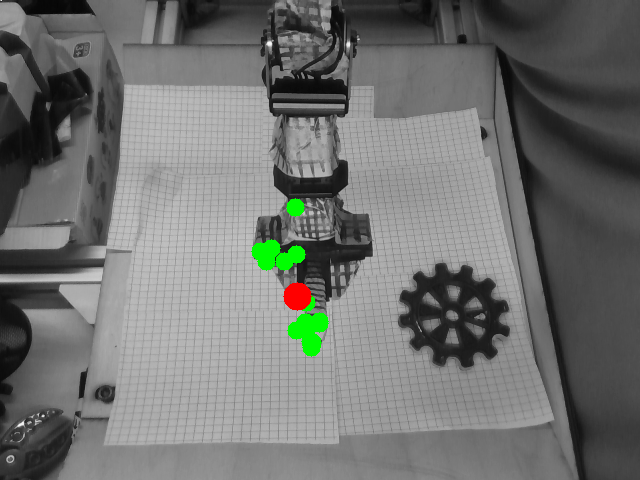} &
    \includegraphics[width=0.17\textwidth]{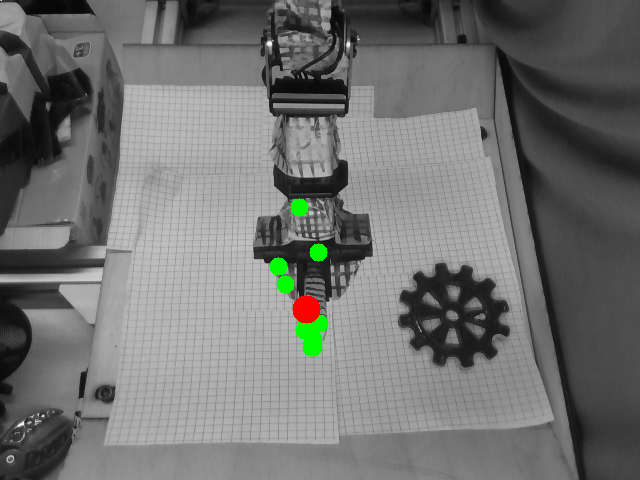} &
    \includegraphics[width=0.17\textwidth]{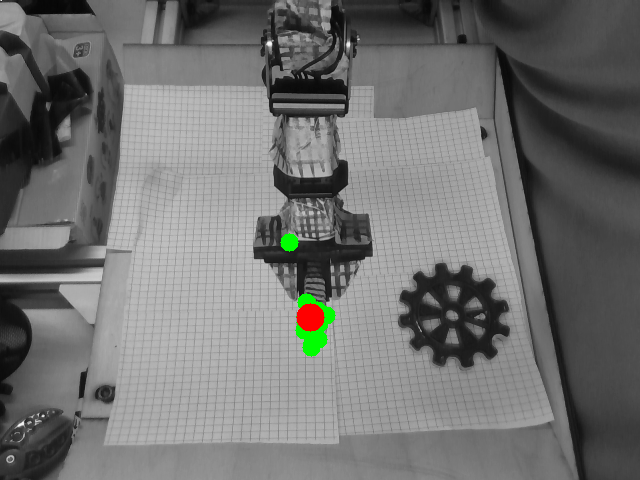} &
    \includegraphics[width=0.17\textwidth]{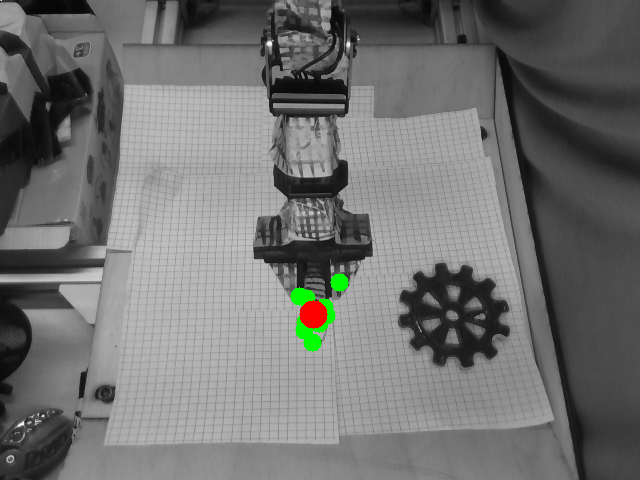}
    \end{tabular}
    \caption{\small{[Best seen in pdf] (Left to right) Original image of the arm with a marker tool, followed by MRCP identification with various values of noise variance. As noise increases, the MRCP points move closer towards the marker tip.}}
    \label{fig:mrcp_noise_egs}
\end{figure*}

The end-effector point identification error plot (Fig~\ref{fig:mrcp_plots}, bottom) provides a clearer comparison of the \method{mavric} ablations,
labeled A through H in the legend. Comparing A and C (1-stage vs. 2-stage search), it is clear that coarse-to-fine MRCP search has a big impact on self-recognition success. Comparing B, C, D, E, and F (increasing noise variance), it is clear that a small amount of noise improves outcomes, but performance deteriorates when the noise is too high. Finally, comparing G, H, and C (top 1 vs top 5 vs top 15 control points), top 15 performs best in most cases. For all remaining experiments, we use variant E (2-stage, noise variance 1.6, and top 15 control point selection). Fig~\ref{fig:mrcp_egs} shows examples of the detected MRCP from various runs under various settings. Fig~\ref{fig:mrcp_noise_egs} shows examples of the effect of noise on MRCP detection with the marker tool, clearly illustrating how higher noise variance biases towards selecting points closer to the tip of the marker. In our experiments, \method{mavric} performed qualitatively well across all these settings,  consistently identifying point close to the tooltips. The end-effector error is also a function of the geometry of the end-effector. For example, in the no-tool case, the parallel jaw gripper has two fingers with a gap in the middle, which is where we annotate the end effector point (see Fig~\ref{fig:body_def}). However, only points on the physical robot can be control points, so this is an additional source of error.

\begin{figure*}
    \centering
    \setlength{\tabcolsep}{0pt}
    \renewcommand{\arraystretch}{1}
    \begin{tabular}{cccc}
    \includegraphics[width=0.27\textwidth]{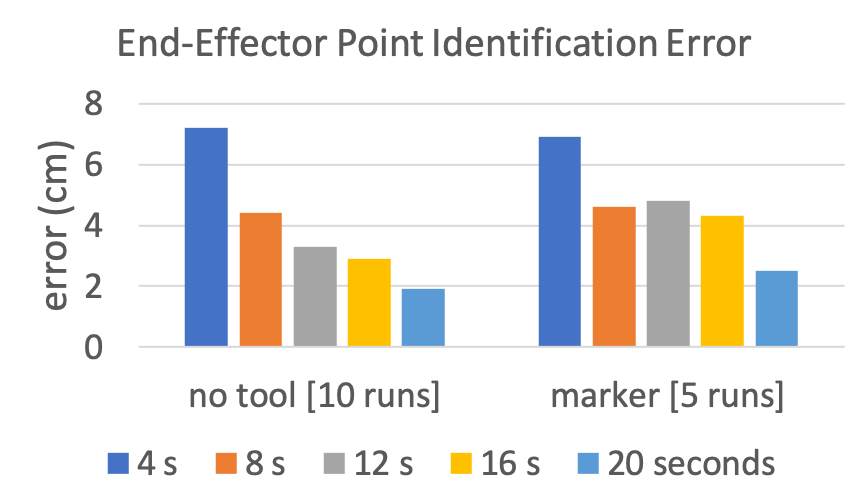} &
    \includegraphics[width=0.27\textwidth]{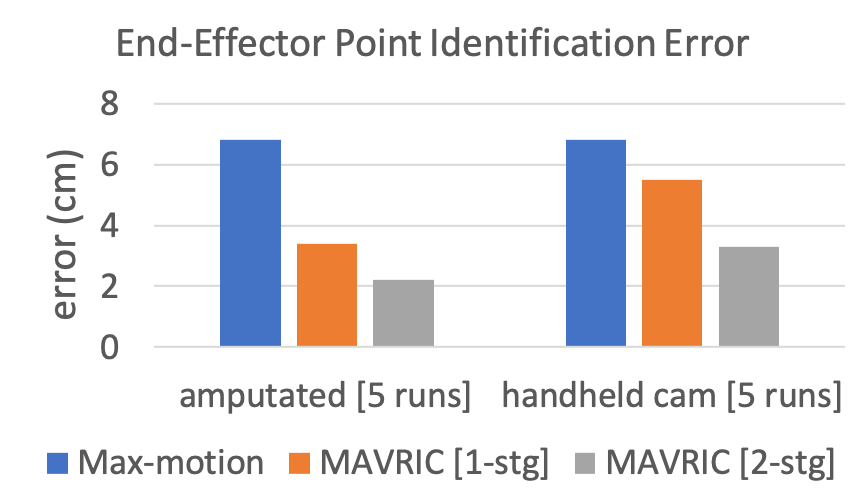} &
    \includegraphics[width=0.22\textwidth]{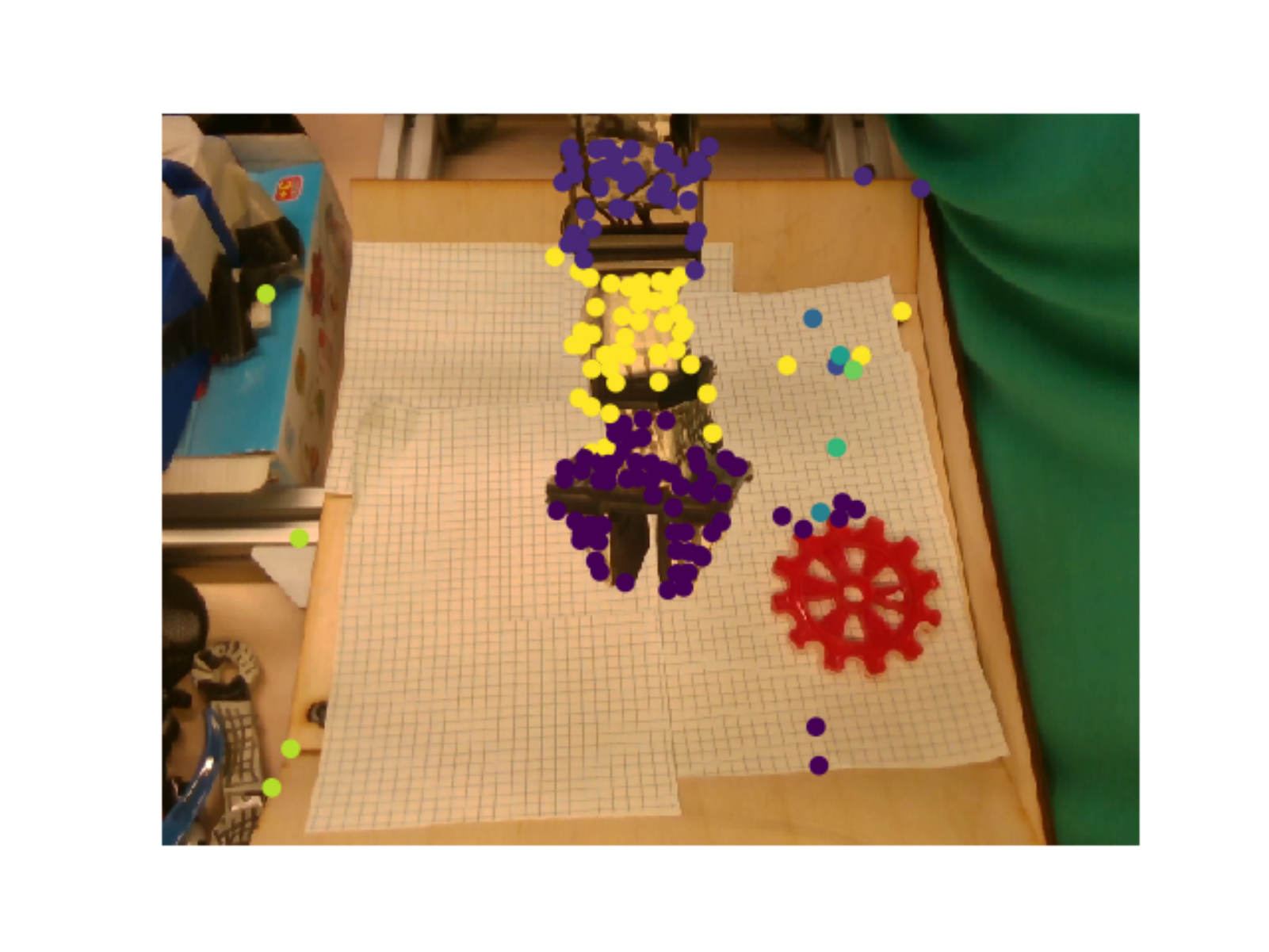} &
    \includegraphics[width=0.22\textwidth]{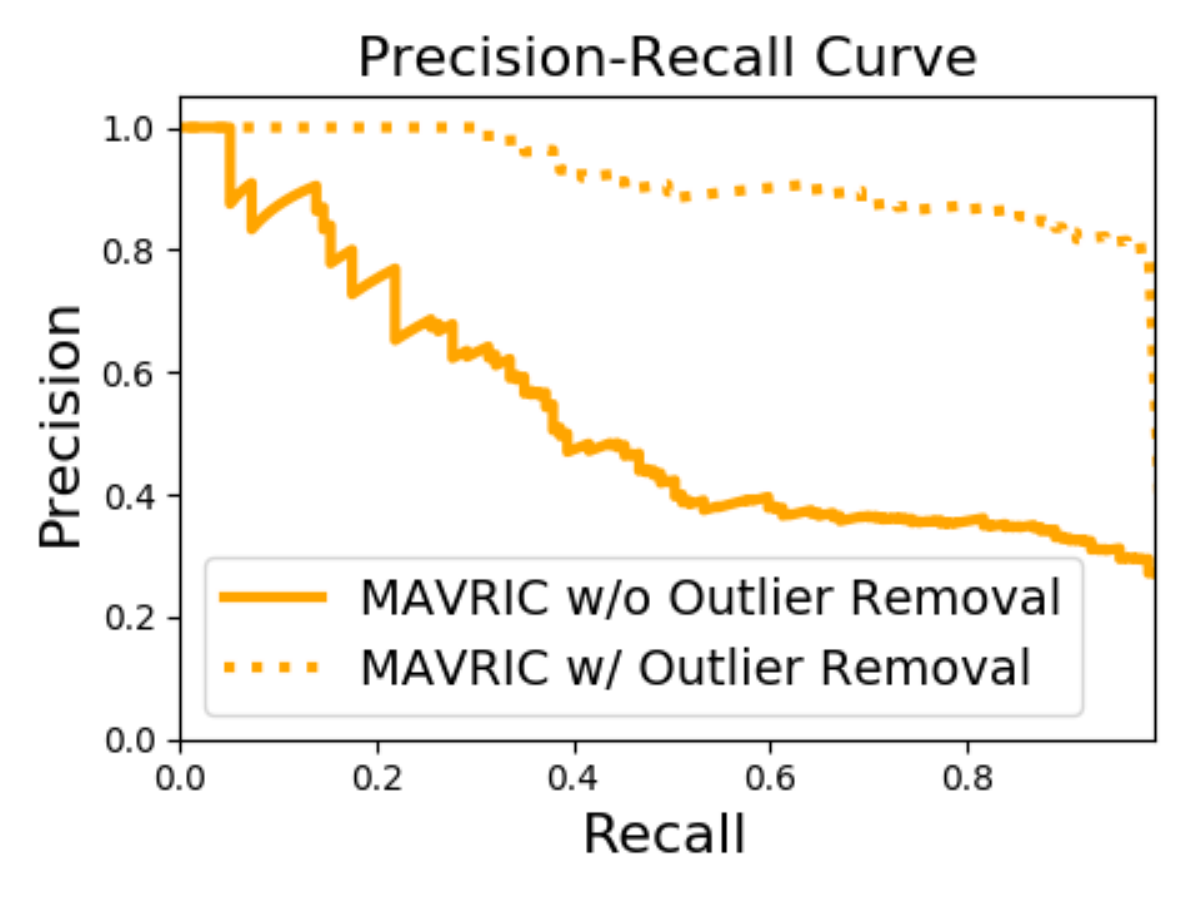}\\
    \small{(a)} & \small(b) & \small(c) & \small(d)
    \end{tabular}
    \caption{\small{[Best seen in pdf] \textbf{(a)} Self-recognition performance with shorter exploration phases, \textbf{(b)} Self-recognition with an amputated arm and a handheld camera,  \textbf{(c)} Discovery of robot arm links from self-recognition phase data: tracks assigned to different clusters are colored differently. \textbf{(d)} Precision-Recall plot for control points.}}
    \label{fig:mrcp_vs_actions}
\end{figure*}

We also quantitatively evaluate self-recognition in two additional settings: an amputated version of the robot arm, with the last two links removed, and a shaky handheld camera. Fig~\ref{fig:mrcp_vs_actions} shows the errors. Once again, 2-stage \method{mavric} works best. Fig~\ref{fig:mrcp_handheld} shows various steps during self-recognition with the handheld camera. Fig~\ref{fig:mrcp_egs} includes an example in the amputated arm setting.

\begin{figure*}
    \centering
    \begin{tabular}{cc}
    video 1, left arm control inputs & video 1, right arm control inputs \\
    \includegraphics[width=0.3\textwidth]{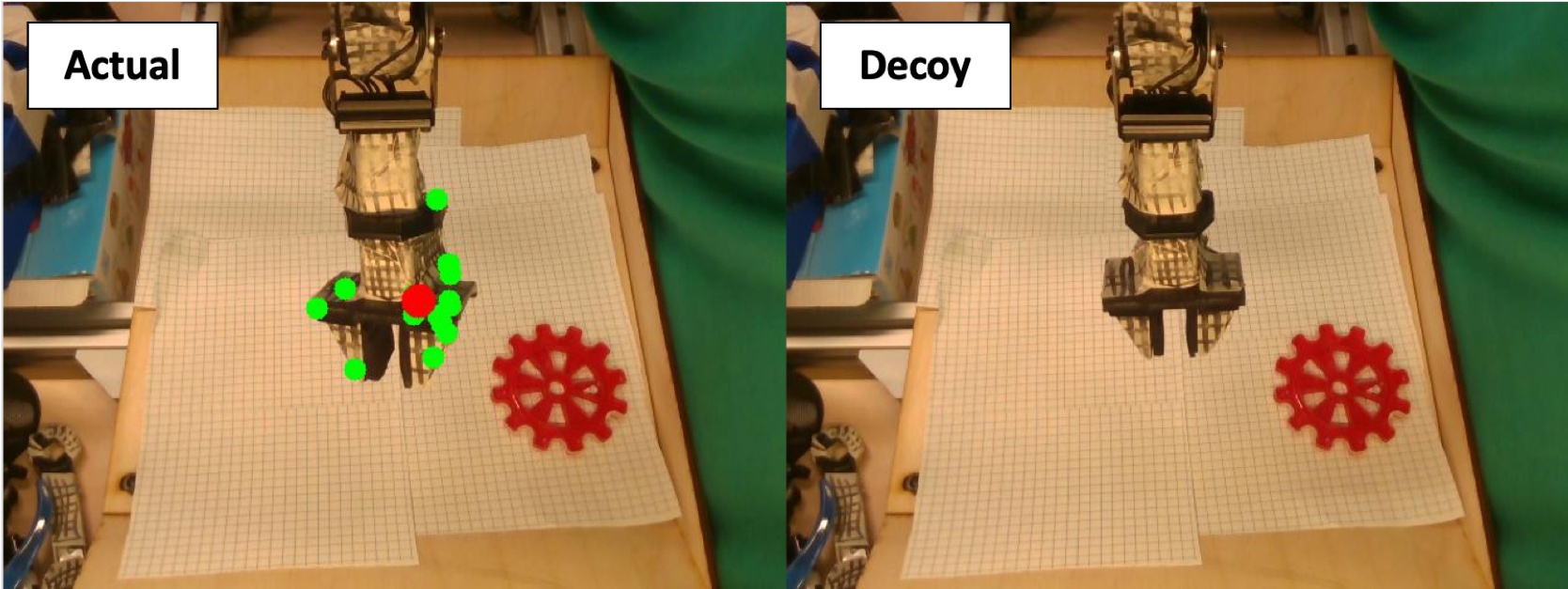} & \includegraphics[width=0.3\textwidth]{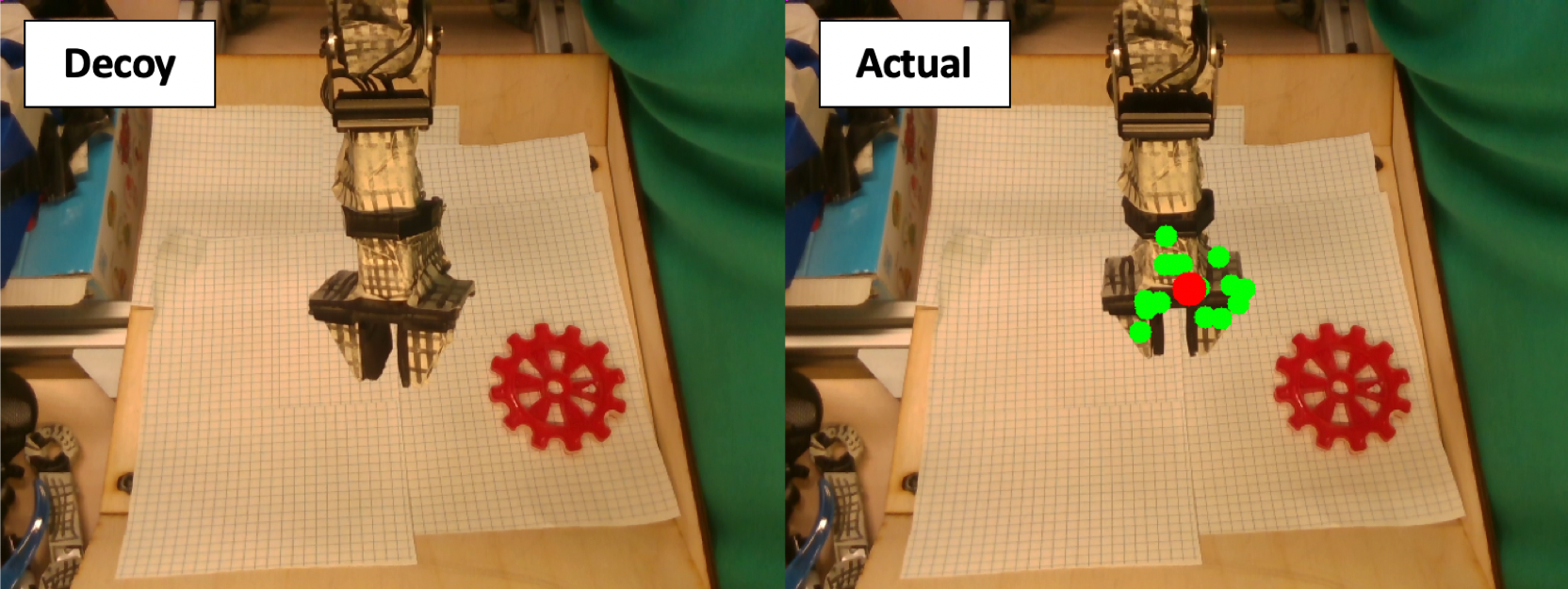}
    \end{tabular}
    \caption{\small{The moving distractors test: given the same video of two arms operating side by side (produced by concatenating individual frames side-by-side from two exploration videos), \method{mavric} correctly ignores the decoy arm and selects the arm that is being controlled based on which control input sequence it receives as input. \noindent\textbf{(left)} \method{mavric} is fed the left arm's controls, and it selects the MRCP (red point) on the left arm's end effector. \noindent\textbf{(right)} With right arm's control inputs, \method{mavric} selects the right arm's end effector.
    }}
    \label{fig:mrcp_side_by_side}
\end{figure*}

\vspace{0.05in} \noindent \textbf{Moving distractors.~~} Next, we evaluate self-recognition with moving distractors by evaluating it on videos with two robots, where one of the robots is controlled by our method, while the other arm, a decoy moves independently, thereby creating a moving distractor.
We create such videos by spatially concatenating two separate exploration videos, so that both arms are moving using the same random motion scheme. Fig~\ref{fig:mrcp_side_by_side} shows an example result. \method{mavric} correctly selects the end-effector of the \emph{correct} arm, based on which arm's control commands it receives as input. See Supp for example videos. Max-motion does not have any control inputs, so it produces the same prediction in both cases.

\vspace{0.05in} \noindent \textbf{Tracking.~~} We now compare against an alternative tracking scheme. Edsinger et al~\cite{edsinger2006can} track moving objects for self-recognition by finding image patches that match the expected appearance of the robot and clustering them based on appearance. We implement their tracking scheme for self-recognition, so that the output is an image patch tracked through the video, representing the end-effector.
On the same ``no-tool'' videos where \method{mavric} correctly identifies the end-effector 10 out of 10 times, this method produces an output image patch that is centered on the end-effector only 2 out of 10 times. Further, since this clustering scheme relies on appearance similarities, it completely breaks down in the moving distractors setting above, where multiple identical-looking robots are present --- the same appearance cluster teleports across the different robots, making responsiveness computation extremely noisy.

\vspace{0.05in} \noindent \textbf{Self-recognition phase duration.~~} While the above results are based on a 100-time step self-recognition phase (approx.~20 s), how much faster could this phase be? We evaluate end-effector identification with even fewer exploration steps in Fig~\ref{fig:mrcp_vs_actions}, which shows that performance deteriorates gracefully under shorter exploration sequences.

\vspace{0.05in} \noindent \textbf{Evaluating control points.~~} While the above results evaluate end-effector identification alone, \method{mavric} finds control points all along the robot body. We now annotate the full robot body to evaluate whether these discovered control points are indeed on the robot body. Treating points on the robot body as ground truth positives, and those outside as negatives, Fig~\ref{fig:mrcp_vs_actions} (d) shows the precision-recall plot as the threshold $\delta$ on the responsiveness scores are varied (``\method{mavric} w/o outlier removal''): while the precision is very high at low recall, it drops off quickly. This is intuitive: the lower the true responsiveness, the more noisy the measurements are. We expect that less responsive control points would benefit from a longer self-recognition phase. However, even with $20$ seconds, it is possible to filter the points to improve the precision-recall performance. We perform simple outlier removal as follows: we measure the 2D position variance of each candidate track over the length of self-recognition, and set a heuristic threshold on this value, below which points are discarded. This simple outlier removal scheme proves sufficient to significantly improve precision-recall, as shown in Fig~\ref{fig:mrcp_vs_actions} (d). These control points may then be clustered based on spatial coherence to discover various links of a rigid robot, and their associated responsiveness scores. Fig~\ref{fig:mrcp_vs_actions} (c) shows an example. We use $K$-means clustering ($K=10$) on position history features.

Please see Supp for videos demonstrating self-recognition in still more varied settings, including different robot configurations, lighting conditions, and more tools.

\begin{table*}[t]
    \centering
    \caption{\small{3D point-reaching error and early termination rate (ETR) during visual servoing for point reaching}}
    \resizebox{1\textwidth}{!}{
    \begin{tabular}{l|cc|cc|cc|cc|cc|cc}
    \toprule
    Method & \multicolumn{2}{c|}{no-tool} & \multicolumn{2}{c|}{wrench} & \multicolumn{2}{c|}{pliers} & \multicolumn{2}{c|}{pencil} & \multicolumn{2}{c|}{marker} & \multicolumn{2}{c}{average}\\
    ~ & error (cm) & ETR (\%) & error (cm) & ETR (\%) & error (cm) & ETR (\%) & error (cm) & ETR (\%) & error (cm) & ETR (\%) & error (cm) & ETR (\%)\\
    \midrule
        ROS MoveIt &$4.4$&-&-&-&-&-&-&-&-&-&-&- \\
        Oracle VS &  $1.2$ & $100$ & $2.3$ & $90$ & $2.2$ & $100$ & $2.2$ & $30$ & $3.6$ & $80$ & $2.3$ & $80$ \\
        \method{mavric}  & $4.4$ & $60$ & $3.5$ & $70$ & $5.7$ & $90$ & $6.8$ & $60$ & $5.9$ & $60$ & $5.2$ & $68$ \\ \bottomrule
    \end{tabular}
    }
    \label{tab:reaching}
\end{table*}

\subsection{Visually Guided Servoing}\label{sec:servo_res}
\new{We now demonstrate controllability using two basic control tasks involving our automatically localized end-effectors: 3D point reaching, and trajectory following. Then, we show how \method{mavric} enables robot-to-robot imitation.}

\noindent \textbf{3D point reaching.~~} We now evaluate \textsc{mavric} (self-recognition + servoing) on 3D-point reaching tasks. %
We set 9 goal positions in the RGBD camera view at varying elevations and azimuths centered at the end effector's initial position at a distance of about 15 cm. We compare \method{mavric} to two methods that have access to additional manually specified information: ``Oracle VS,'' which servos a manually annotated end effector using the same visual servoing approach (based on~\cite{jagersand1997experimental,bonkovi2008population}) as our method, and ``ROS MoveIt''~\cite{moveit} which has knowledge of the full robot morphology and kinematics models, camera calibration matrices, and proprioception. %
We allow a maximum of 150 steps.%

Tab~\ref{tab:reaching} reports (i) median 3D distance error of the manually annotated end effector point from the goal, and (ii) early termination rate (ETR). Early termination is triggered whenever the MRCP has reached within a 5 px radius of the goal --- in our experiments, this is a good proxy for servoing success. ROS MoveIt cannot control unmodeled tools, so we report its performance only in the no-tool setting. Its error is higher than Oracle VS; this may be due to WidowX robot model inaccuracy, servo encoder position errors, and camera calibration error. Oracle VS does well in most settings, and \method{mavric} takes slightly longer (lower ETR), but its error is within 3 cm of Oracle VS --- in our experiments, this is largely explained by the end-effector point identification error (Fig~\ref{fig:mrcp_plots}) from the self-recognition stage. We use standard techniques for visual servoing (Sec~\ref{sec:servoing}), and Oracle VS and \method{mavric} both inherit common problems of these methods, such as local minima and singularites during Jacobian estimation. This sometimes results in failures with unbounded errors --- we report the median error to discard these.  %

\begin{figure}%
    \centering
    \includegraphics[width=0.35\textwidth]{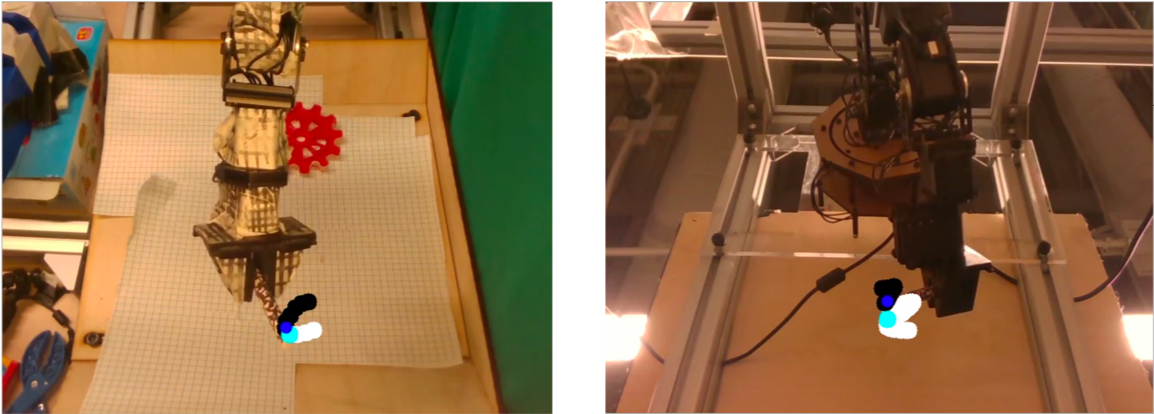}
    \caption{\small{[Best seen in pdf] Robot-to-robot imitation with \method{mavric}. In both video frames, cyan represents the next target, white points are future targets, and black points are previously reached targets. \textbf{(left)} A video frame of a source robot draws the letter C --- in this case, we used \method{mavric} to perform this task with visual servoing for trajectory-following. \textbf{(right)} A video frame of the target robot imitating the motions of the source robot. Full video in Supp.}}
    \label{fig:imitation}
\end{figure}

\vspace{0.05in} \noindent \textbf{Trajectory following.~~} Aside from single point reaching, we may also servo to follow a trajectory. In Supp, we show videos demonstrating \method{mavric} writing the letter ``C'' onto the projected view of the camera through 2D trajectory following, with different robots. Fig~\ref{fig:imitation} (left) shows one such video frame.

\vspace{0.05in} \noindent \textbf{Robot-to-robot imitation.~~} Consider a source robot with unknown morphology and kinematics that is performing a writing task --- perhaps it has been trained for days using an RL approach~\cite{levine2018learning}. How might a target robot, also with unknown morphology and kinematics, perform the same task? This requires visually mapping the embodiment across these two robots, with very little data --- we know of no prior approach that might accomplish this. Using \method{mavric}, we map the automatically discovered MRCP of the source robot to that of the target, and perform imitation by servoing the target MRCP along the observed trajectory of the source MRCP. %
Fig~\ref{fig:imitation} shows frames from the result videos (full videos in Supp). Here, the source and target robots are both WidowX arms, with different robot and camera poses, and different illumination.
\section{Discussion}
We have presented \method{mavric}, an approach that performs fast robot body recognition and uses this to accomplish morphology-agnostic visuomotor control. Reinforcement learning-based approaches operating in the same setting typically require days of robot data~\cite{finn2016deep,sadeghi2017sim2real,levine2018learning,james2019sim,levine2018learning,ebert2017self}, compared to 20s for \method{mavric} --- we know of no other methods that can perform visuomotor control without knowledge of kinematics or morphology on similar timescales to \method{mavric}. We have demonstrated \method{mavric} across diverse settings, with various tools, shaky camera views etc. %
\new{However, our validation of \method{mavric} has been limited to coarse control for basic tasks due to various limitations: end-effector self-recognition error (average ~2 cm), point tracking failures under occlusions, noisy depth sensing causing poor Jacobian estimates, and well-known issues with visual servoing~\cite{chaumette1998potential,hutchinson1996tutorial,cai20176d,cho2019sampling}. We hope to address these in future work by moving to multi-camera setups, using longer exploration phases, using better tracking algorithms, and better depth sensors. \method{mavric} may also be extended to handle settings with multiple end-effectors such as a multi-fingered hand.}
\IEEEpeerreviewmaketitle

\bibliographystyle{IEEEtran}
\bibliography{references}

\begin{thebibliography}{10}
\providecommand{\url}[1]{#1}
\csname url@rmstyle\endcsname
\providecommand{\newblock}{\relax}
\providecommand{\bibinfo}[2]{#2}
\providecommand\BIBentrySTDinterwordspacing{\spaceskip=0pt\relax}
\providecommand\BIBentryALTinterwordstretchfactor{4}
\providecommand\BIBentryALTinterwordspacing{\spaceskip=\fontdimen2\font plus
\BIBentryALTinterwordstretchfactor\fontdimen3\font minus
  \fontdimen4\font\relax}
\providecommand\BIBforeignlanguage[2]{{%
\expandafter\ifx\csname l@#1\endcsname\relax
\typeout{** WARNING: IEEEtran.bst: No hyphenation pattern has been}%
\typeout{** loaded for the language `#1'. Using the pattern for}%
\typeout{** the default language instead.}%
\else
\language=\csname l@#1\endcsname
\fi
#2}}

\bibitem{gallup1970chimpanzees}
G.~G. Gallup, ``Chimpanzees: self-recognition,'' \emph{Science}, vol. 167, no.
  3914, pp. 86--87, 1970.

\bibitem{asendorpf1996self}
J.~B. Asendorpf, V.~Warkentin, and P.-M. Baudonniere, ``Self-awareness and
  other-awareness.'' \emph{Developmental Psychology}, 1996.

\bibitem{shannon1948mathematical}
C.~E. Shannon, ``A mathematical theory of communication,'' \emph{Bell system
  technical journal}, vol.~27, no.~3, pp. 379--423, 1948.

\bibitem{lucas1981iterative}
B.~D. Lucas, T.~Kanade, \emph{et~al.}, ``An iterative image registration
  technique with an application to stereo vision,'' 1981.

\bibitem{corke1993visual}
P.~I. Corke, ``Visual control of robot manipulators--a review,'' in
  \emph{Visual Servoing: Real-Time Control of Robot Manipulators Based on
  Visual Sensory Feedback}, 1993.

\bibitem{hutchinson1996tutorial}
S.~Hutchinson, G.~D. Hager, and P.~I. Corke, ``A tutorial on visual servo
  control,'' \emph{IEEE transactions on robotics and automation}, vol.~12,
  no.~5, pp. 651--670, 1996.

\bibitem{chaumette2006visual}
F.~Chaumette and S.~Hutchinson, ``{Visual servo control. I. Basic
  approaches},'' \emph{IEEE Robotics \& Automation Magazine}, 2006.

\bibitem{chaumette2007visual}
------, ``Visual servo control, part ii: Advanced approaches,'' \emph{IEEE
  Robotics and Automation Magazine}, vol.~14, no.~1, pp. 109--118, 2007.

\bibitem{marchand2005visp}
{\'E}.~Marchand, F.~Spindler, and F.~Chaumette, ``{ViSP} for visual servoing,''
  \emph{IEEE Robotics \& Automation Magazine}, 2005.

\bibitem{hosoda1994versatile}
K.~Hosoda and M.~Asada, ``Versatile visual servoing without knowledge of true
  jacobian,'' in \emph{IROS}.\hskip 1em plus 0.5em minus 0.4em\relax IEEE,
  1994.

\bibitem{yoshimi1994active}
B.~H. Yoshimi and P.~K. Allen, ``Active, uncalibrated visual servoing,'' in
  \emph{ICRA}.\hskip 1em plus 0.5em minus 0.4em\relax IEEE, 1994.

\bibitem{jagersand1997experimental}
M.~Jagersand, O.~Fuentes, and R.~Nelson, ``Experimental evaluation of
  uncalibrated visual servoing for precision manipulation,'' in \emph{ICRA},
  1997.

\bibitem{bonkovi2008population}
M.~Bonkovi, K.~Jezernik, \emph{et~al.}, ``Population-based uncalibrated visual
  servoing,'' \emph{IEEE/ASME Transactions on Mechatronics}, 2008.

\bibitem{shademan2010robust}
A.~Shademan, A.-M. Farahmand, and M.~J{\"a}gersand, ``Robust jacobian
  estimation for uncalibrated visual servoing,'' in \emph{ICRA}, 2010.

\bibitem{dame2011mutual}
A.~Dame and E.~Marchand, ``Mutual information-based visual servoing,''
  \emph{IEEE Transactions on Robotics}, vol.~27, no.~5, pp. 958--969, 2011.

\bibitem{lampe2013acquiring}
T.~Lampe and M.~Riedmiller, ``Acquiring visual servoing reaching and grasping
  skills using neural reinforcement learning,'' in \emph{IJCNN}, 2013.

\bibitem{mohta2014vision}
K.~Mohta, V.~Kumar, and K.~Daniilidis, ``Vision-based control of a quadrotor
  for perching on lines,'' in \emph{ICRA}, 2014.

\bibitem{bateux2018training}
Q.~Bateux, E.~Marchand, J.~Leitner, F.~Chaumette, and P.~Corke, ``Training deep
  neural networks for visual servoing,'' in \emph{ICRA}, 2018.

\bibitem{viereck2017learning}
U.~Viereck, A.~t. Pas, K.~Saenko, and R.~Platt, ``Learning a visuomotor
  controller for real world robotic grasping using simulated depth images,''
  \emph{arXiv preprint arXiv:1706.04652}, 2017.

\bibitem{andersson1988robot}
R.~L. Andersson, ``A robot ping-pong player,'' \emph{Experiment in Real-Time
  Intelligent Control.}, 1988.

\bibitem{jang1991concepts}
W.~Jang, K.~Kim, M.~Chung, and Z.~Bien, ``Concepts of augmented image space and
  transformed feature space for efficient visual servoing of an “eye-in-hand
  robot”,'' \emph{Robotica}, 1991.

\bibitem{cai20146d}
C.~Cai, E.~Dean-Le{\'o}n, N.~Somani, and A.~Knoll, ``6d image-based visual
  servoing for robot manipulators with uncalibrated stereo cameras,'' in
  \emph{IROS}, 2014.

\bibitem{mcfadyen2017image}
A.~McFadyen, M.~Jabeur, and P.~Corke, ``Image-based visual servoing with
  unknown point feature correspondence.'' \emph{RAL}, 2017.

\bibitem{finn2016deep}
C.~Finn, X.~Y. Tan, Y.~Duan, T.~Darrell, S.~Levine, and P.~Abbeel, ``Deep
  spatial autoencoders for visuomotor learning,'' in \emph{ICRA}.\hskip 1em
  plus 0.5em minus 0.4em\relax IEEE.

\bibitem{ebert2017self}
F.~Ebert, C.~Finn, A.~X. Lee, and S.~Levine, ``Self-supervised visual planning
  with temporal skip connections,'' \emph{arXiv preprint arXiv:1710.05268},
  2017.

\bibitem{sadeghi2017sim2real}
F.~Sadeghi, A.~Toshev, E.~Jang, and S.~Levine, ``Sim2real view invariant visual
  servoing by recurrent control,'' \emph{arXiv preprint arXiv:1712.07642},
  2017.

\bibitem{levine2018learning}
S.~Levine, P.~Pastor, A.~Krizhevsky, J.~Ibarz, and D.~Quillen, ``Learning
  hand-eye coordination for robotic grasping with deep learning and large-scale
  data collection,'' \emph{IJRR}, 2018.

\bibitem{lee2017learning}
A.~X. Lee, S.~Levine, and P.~Abbeel, ``Learning visual servoing with deep
  features and fitted q-iteration,'' \emph{arXiv:1703.11000}, 2017.

\bibitem{james2019sim}
S.~James, P.~Wohlhart, M.~Kalakrishnan, D.~Kalashnikov, A.~Irpan, J.~Ibarz,
  S.~Levine, R.~Hadsell, and K.~Bousmalis, ``Sim-to-real via sim-to-sim:
  Data-efficient robotic grasping via randomized-to-canonical adaptation
  networks,'' in \emph{CVPR}, 2019.

\bibitem{shkurti2017underwater}
F.~Shkurti, W.-D. Chang, P.~Henderson, M.~J. Islam, J.~C.~G. Higuera, J.~Li,
  T.~Manderson, A.~Xu, G.~Dudek, and J.~Sattar, ``Underwater multi-robot
  convoying using visual tracking by detection,'' in \emph{IROS}, 2017.

\bibitem{islam2018towards}
M.~J. Islam, M.~Fulton, and J.~Sattar, ``Towards a generic diver-following
  algorithm: Balancing robustness and efficiency in deep visual detection,''
  \emph{arXiv preprint arXiv:1809.06849}, 2018.

\bibitem{michel2004motion}
P.~Michel, K.~Gold, and B.~Scassellati, ``Motion-based robotic
  self-recognition,'' in \emph{IROS}, 2004.

\bibitem{natale2007sensorimotor}
L.~Natale, F.~Orabona, G.~Metta, and G.~Sandini, ``Sensorimotor coordination in
  a “baby” robot: learning about objects through grasping,'' \emph{Progress
  in brain research}, vol. 164, pp. 403--424, 2007.

\bibitem{jonschkowski2018found}
R.~Jonschkowski, ``{Found by NEMO: Unsupervised Object Detection from Negative
  Examples and Motion},'' \emph{OpenReview ICLR submission}, 2018.

\bibitem{byravan2018se3}
A.~Byravan, F.~Lceb, F.~Meier, and D.~Fox, ``{SE3-Pose-Nets}: Structured deep
  dynamics models for visuomotor control,'' in \emph{ICRA}, 2018.

\bibitem{edsinger2006can}
A.~Edsinger and C.~C. Kemp, ``What can i control? a framework for robot
  self-discovery,'' in \emph{6th International Conference on Epigenetic
  Robotics}, 2006.

\bibitem{cover2012elements}
T.~M. Cover and J.~A. Thomas, \emph{Elements of information theory}.\hskip 1em
  plus 0.5em minus 0.4em\relax John Wiley \& Sons, 2012.

\bibitem{choi2018contingencyaware}
J.~Choi, Y.~Guo, M.~Moczulski, J.~Oh, N.~Wu, M.~Norouzi, and H.~Lee,
  ``Contingency-aware exploration in reinforcement learning,'' in \emph{ICLR},
  2019.

\bibitem{kraskov2004estimating}
A.~Kraskov, H.~St{\"o}gbauer, and P.~Grassberger, ``Estimating mutual
  information,'' \emph{Physical review E}, vol.~69, no.~6, p. 066138, 2004.

\bibitem{ver2000non}
G.~Ver~Steeg, ``Non-parametric entropy estimation toolbox (npeet),'' 2000.

\bibitem{shi1993good}
J.~Shi and C.~Tomasi, ``Good features to track,'' Cornell University, Tech.
  Rep., 1993.

\bibitem{yan2006general}
J.~Yan and M.~Pollefeys, ``A general framework for motion segmentation:
  Independent, articulated, rigid, non-rigid, degenerate and non-degenerate,''
  in \emph{ECCV}, 2006.

\bibitem{yang2019replab}
B.~Yang, J.~Zhang, D.~Jayaraman, and S.~Levine, ``Replab: A reproducible
  low-cost arm benchmark platform for robotic learning,'' \emph{ICRA}, 2019.

\bibitem{yang2019replab2}
B.~Yang, J.~Zhang, V.~Pong, S.~Levine, and D.~Jayaraman, ``{REPLAB: A
  Reproducible Low-Cost Arm Benchmark Platform for Robotic Learning},''
  \emph{arXiv preprint arXiv:1905.07447}, 2019.

\bibitem{moveit}
``{ROS MoveIt} package,'' \url{https://moveit.ros.org/}, accessed: 2019-07-06.

\bibitem{chaumette1998potential}
F.~Chaumette, ``Potential problems of stability and convergence in image-based
  and position-based visual servoing,'' in \emph{The confluence of vision and
  control}.\hskip 1em plus 0.5em minus 0.4em\relax Springer, 1998, pp. 66--78.

\bibitem{cai20176d}
C.~Cai, ``6d visual servoing for industrial manipulators applied to human-robot
  interaction scenarios,'' 2017.

\bibitem{cho2019sampling}
S.~Cho and D.~H. Shim, ``Sampling-based visual path planning framework for a
  multirotor uav,'' \emph{International Journal of Aeronautical and Space
  Sciences}, pp. 1--29, 2019.

\end{thebibliography}

\end{document}